\documentclass[11pt]{article}

\PassOptionsToPackage{table}{xcolor}
\usepackage[final]{acl}
\usepackage{enumitem}
\usepackage{times}
\usepackage{latexsym}
\usepackage{booktabs}
\usepackage{colortbl}
\definecolor{darkblue}{rgb}{0, 0, 0.5}
\usepackage{multirow}
\usepackage{amsmath}
\usepackage{amssymb}
\usepackage[T1]{fontenc}
\usepackage[utf8]{inputenc}

\usepackage{microtype}

\usepackage{inconsolata}

\usepackage{graphicx}
\graphicspath{{../}{./}}

\usepackage[most]{tcolorbox}

\definecolor{lightgray}{gray}{0.93}
\definecolor{lightblue}{RGB}{220,235,247}
\newcommand{\sert}[1]{\cellcolor{lightblue}#1}
\newcommand{\pos}[1]{\textcolor{red}{+{#1}}}
\newcommand{\negcolor}[1]{\textcolor{blue}{#1}}

\newtcolorbox{bluebox}[1][]{
  enhanced,
  colframe=blue!40!gray,
  colback=white,
  coltitle=white,
  colbacktitle=blue!40!gray,
  width=\linewidth,
  arc=2mm,
  auto outer arc,
  boxrule=0.5pt,
  left=10pt,
  right=10pt,
  drop shadow={black!50!white},
  top=10pt,
  bottom=10pt,
  title={#1},
  fonttitle=\bfseries,
  title code={\node[rounded corners, fill=blue!75!black, draw=none, text=white] at (frame.title) {\textbf{#1}};},
  attach boxed title to top center={yshift=-2mm},
  boxed title style={sharp corners, size=small}
}

\title{Too Correct to Learn: Reinforcement Learning on Saturated Reasoning Data}

\author{
Zhenwen Liang$^{1,\dagger}$,
Yujun Zhou$^{1,2,}$$^\dagger$\thanks{Work done during Yujun's Internship at Tencent AI Lab.}\;,
\textbf{
Sidi Lu$^1$,
Xiangliang Zhang$^2$,
Haitao Mi$^1$,
Dong Yu$^1$}
\vspace{1em}\\
$^1$Tencent AI Lab,
$^2$University of Notre Dame,
\vspace{0.1cm} \\ \vspace{0.1cm}
$^\dagger$ Equal contribution \quad
\\
Correspondence to:  \texttt{zhenwzliang@global.tencent.com}
}

\begin{document}
\maketitle
\begin{abstract}
Reinforcement Learning (RL) enhances LLM reasoning, yet a paradox emerges as models scale: strong base models saturate standard benchmarks (e.g., MATH), yielding correct but homogeneous solutions. In such environments, the lack of failure cases causes the advantage signal in group-relative algorithms (e.g., GRPO) to vanish, driving policies into mode collapse. To address this, we propose Constrained Uniform Top-K Sampling (CUTS), a parameter-free decoding strategy enforcing structure-preserving exploration. Unlike standard sampling that follows model biases, CUTS flattens the local optimization landscape by sampling uniformly from constrained high-confidence candidates. We integrate this into Mixed-CUTS, a training framework synergizing exploitative and exploratory rollouts to amplify intra-group advantage variance. Experiments on Qwen3 models demonstrate that our approach prevents policy degeneration and significantly boosts out-of-domain generalization. Notably, Mixed-CUTS improves Pass@1 accuracy on the challenging AIME25 benchmark by up to 15.1\% over standard GRPO, validating that maintaining diversity within the high-probability region of the model distribution is critical for rigorous reasoning.
\end{abstract}

\section{Introduction}

RL is central to aligning Large Language Models (LLMs) with complex reasoning tasks \citep{jaech2024openai,guo2025deepseek,yang2025qwen3}. Leveraging outcome-based supervision, algorithms like Group Relative Policy Optimization (GRPO) \citep{shao2024deepseekmath} transform models from pattern matchers into rigorous reasoners \citep{yu2025dapo,zheng2025parallel,huang2025r,liang2025clue,liu2025vogue,liu2025stable,zhao2025one,zhou2026capability}.

However, RL's efficacy increasingly hinges on data difficulty. High-quality reasoning datasets are scarce and rapidly absorbed into community training pipelines, rendering benchmarks \textit{saturated}—where strong base models already solve most instances \citep{liu2025deepseek,yang2025qwen3}. This growing prevalence of \textbf{saturated reasoning data} fundamentally alters the learning dynamics of RL for LLMs.

For strong base models (e.g., Qwen3), standard datasets like MATH have saturated, yielding high baseline success rates \citep{yang2025qwen3}. This poses a critical challenge for group-relative learning: when a model generates homogeneous correct solutions, intra-group reward variance collapses toward zero. Lacking failure cases or contrast, the relative advantage signal vanishes \citep{zhu2025surprising}. Consequently, the model succumbs to \textbf{saturation-induced mode collapse}—not because it is incorrect, but because it is \emph{too correct to learn}. The policy becomes trapped in local optima of ``easy successes,'' ceasing to explore generalizable strategies \citep{zhou2025evolving}. Standard entropy regularization fails here by indiscriminately penalizing confidence, disrupting coherent reasoning rather than restoring learning signals \citep{cui2025entropy}.

This reveals a structural limitation: on saturated data, correctness alone provides insufficient training signals. To address this, we argue that effective RL requires explicitly reintroducing diversity to reignite the advantage signal. Rather than altering objectives, we focus on the decoding process as a controllable intervention point.

We introduce \textbf{Constrained Uniform Top-K Sampling (CUTS)}, an inference-time operator designed to break the ``rich-get-richer'' dynamics of standard sampling. Instead of adhering to skewed distributions that favor dominant paths, CUTS flattens the local landscape by sampling uniformly from a constrained set of high-confidence (Top-$K$) candidates. This decouples generation probability from historical preference, compelling the model to explore semantically valid but underestimated tokens. By restricting uniformity to this confidence-filtered window, CUTS ensures structural coherence while enabling controlled exploration.

We integrate this operator into \textbf{Mixed-CUTS}, a framework leveraging a dual-stream rollout strategy to amplify GRPO's intra-group variance. For each prompt, we generate a mixture of \textit{exploitative} (standard sampling) and \textit{exploratory} (CUTS) trajectories. This hybrid design anchors the baseline while injecting necessary diversity. Crucially, even when all solutions are correct, the structural contrast between standard and CUTS-induced paths restores informative gradient signals, preventing convergence stagnation on saturated benchmarks.

\textbf{Contributions.} 
(1) We diagnose and formalize saturation-induced collapse: a failure mode in group-relative RL where high baseline correctness on easy datasets causes the advantage signal to vanish. 
(2) We propose Mixed Constrained Uniform Top-K Sampling (Mixed-CUTS), a parameter-free decoding operator that enforces structure-preserving exploration to counteract this stagnation. 
(3) Empirically, we demonstrate significant generalization gains on various benchmarks, validating that maintaining diversity is essential when correctness alone provides insufficient signal.

% \vspace{-1mm}
\section{Method}
\label{sec:method}
% \vspace{-1mm}
\subsection{Preliminaries}
% \vspace{-1mm}
We build our training framework upon GRPO \citep{shao2024deepseekmath}. Unlike PPO \citep{schulman2017proximal} that requires a parametric value network, GRPO eliminates the critic to reduce memory overhead, instead using group statistics as a baseline. Formally, given query $\mathbf{q}$, the reference policy $\pi_{\theta_{\text{old}}}$ samples $G$ outputs $\{\mathbf o_1,\ldots,\mathbf o_G\}$, yielding rewards $\{r_1,\ldots,r_G\}$. Advantages are computed by standardizing rewards within the group:
\vspace{-2.7mm}
\begin{equation}
\label{eq:advantage}
\small
\hat{A}_i=\frac{r_i-\text{mean}(r_1,\ldots,r_G)}{\text{std}(r_1,\ldots,r_G)+\epsilon}
\end{equation}
\par\vspace{-2.7mm}
Crucially, this \emph{trajectory-level} advantage is applied uniformly to every token $t$, i.e., $\hat{A}_{i,t}=\hat{A}_i$. The policy $\theta$ is then updated by maximizing a clipped surrogate objective that ensures stability within a trust region:
\vspace{-2.7mm}
\begin{equation}
\small
\begin{aligned}
&\frac{1}{G}\sum_{i=1}^{G}\frac{1}{|o_i|}\sum_{t=1}^{|o_i|}
\min\!\Biggl\{
\frac{\pi_{\theta}\!\big(o_{i,t}\mid \mathbf{q},\mathbf{o}_{i,<t}\big)}
     {\pi_{\theta_{\mathrm{old}}}\!\big(o_{i,t}\mid \mathbf{q},\mathbf{o}_{i,<t}\big)}
     \,\hat{A}_{i,t},
\\
&\operatorname{clip}\!\left(
\frac{\pi_{\theta}\!\big(o_{i,t}\mid \mathbf{q},\mathbf{o}_{i,<t}\big)}
     {\pi_{\theta_{\mathrm{old}}}\!\big(o_{i,t}\mid \mathbf{q},\mathbf{o}_{i,<t}\big)},
\,1-\epsilon_{\mathrm{low}},\,1+\epsilon_{\mathrm{high}}
\right)\hat{A}_{i,t}
\Biggr\}
\end{aligned}
\end{equation}

\textbf{The Vanishing-Advantage Problem.}
A critical limitation of Eq.~\ref{eq:advantage} arises on \emph{saturated} datasets. If the model succeeds on all $G$ trajectories (i.e., $r_i = 1, \forall i$), the standard deviation $\text{std}(r)$ becomes zero, causing the standardized advantage $\hat{A}_i$ to vanish or depend solely on the stabilizer $\epsilon$. Consequently, optimization stalls despite high accuracy, as the lack of contrast eliminates the learning signal. This necessitates a mechanism to explicitly guarantee non-zero intra-group variance.

\begin{figure*}[t]
\centering
\includegraphics[width=0.88\textwidth]{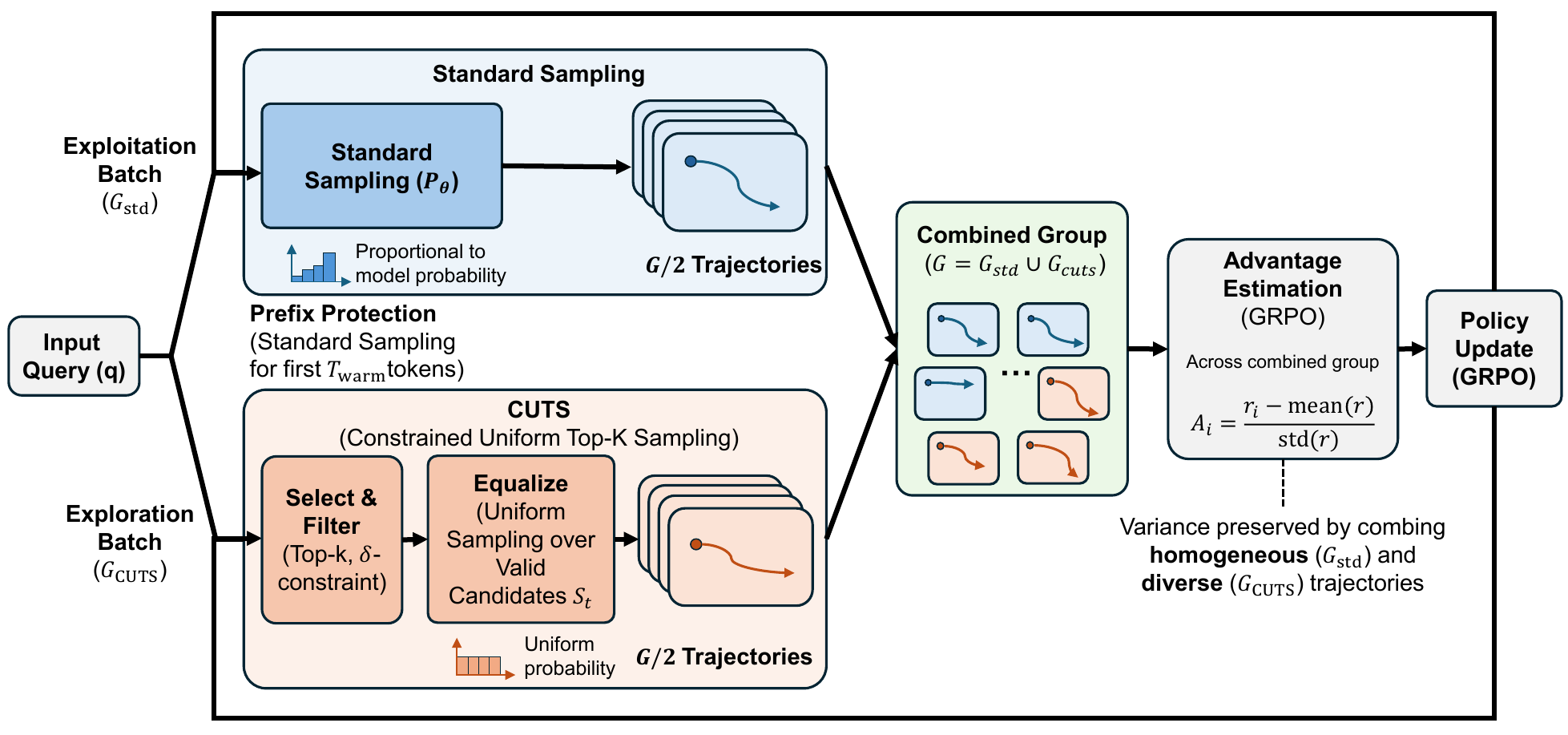}
\caption{\textbf{The Mixed-CUTS Framework.} The framework combines exploitative rollouts ($\mathcal{G}_\text{std}$) and exploratory rollouts ($\mathcal{G}_\text{CUTS}$) to preserve advantage variance under saturated training conditions. The CUTS operator enforces uniform sampling within a constrained Top-$K$ candidate set, decoupling generation from model bias.}
\label{fig:framework}
\vspace{-4mm}
\end{figure*}

\subsection{Constrained Uniform Top-K Sampling (CUTS)}

Standard autoregressive decoding samples $x_t$ from $P_\theta(x_t \mid \mathbf{q}, \mathbf{x}_{<t})$. On saturated data, however, this proportional nature induces \textbf{mode collapse}: distributions become excessively peaked around dominant paths, suppressing valid alternatives. To counteract this, we propose \textbf{Constrained Uniform Top-K Sampling (CUTS)}, a parameter-free operator that constructs a locally flattened proposal distribution $Q(x_t \mid \mathbf{q}, \mathbf{x}_{<t})$ via a three-stage process: \textit{Select, Filter, and Equalize}. CUTS introduces no additional trainable parameters and operates purely at inference time, with a small set of decoding hyperparameters.

\vspace{-2mm}\paragraph{Selection and Filtering.}
At step $t$, we extract the top-$K$ tokens $\mathcal{V}_{\text{top-}K}$. To preclude incoherent tail generations, we apply a probability threshold $\delta$ to define the valid candidate set:
\vspace{-2mm}
\begin{equation}
\mathcal{S}_t = \left\{ v \in \mathcal{V}_{\text{top-}K} \mid P_\theta(v \mid \mathbf{q}, \mathbf{x}_{<t}) \ge \delta \right\}.
\end{equation}
\par\vspace{-2mm}
\noindent This constraint restricts exploration to a high-confidence token neighborhood, serving as a proxy for local plausibility. We handle edge cases explicitly: if $\mathcal{S}_t=\varnothing$, we fallback to $\mathcal{V}_{\text{top-}K}$; if $|\mathcal{S}_t|=1$, the operator degenerates to a deterministic choice.

\vspace{-2.5mm}\paragraph{Uniform Equalization.}
To decouple generation probability from the model's bias within $\mathcal{S}_t$, we define the proposal distribution as a uniform prior:
\vspace{-2mm}
\begin{equation}
Q(x_t = v \mid \mathbf{q}, \mathbf{x}_{<t}) =
\begin{cases}
\frac{1}{|\mathcal{S}_t|} & \text{if } v \in \mathcal{S}_t, \\
0 & \text{otherwise}.
\end{cases}
\end{equation}
\par\vspace{-2mm}
\noindent 
This equalization enforces local ``width-first'' exploration, enabling the traversal of plausible reasoning paths that are underrepresented in the standard distribution.

\vspace{-2.5mm}\paragraph{Prefix Protection.} Given the sensitivity of reasoning tasks to early decisions, we employ a warm-up mechanism to preserve initial stability. We apply CUTS only after a prefix of $T_{\text{warm}}$ tokens; prior to this, standard sampling is used to establish a coherent problem setup.

\subsection{The Mixed-CUTS Training Framework}To balance exploration with policy stability, we introduce the \textbf{Mixed-CUTS} framework within the GRPO paradigm in Figure \ref{fig:framework}. For each query $\mathbf{q}$, we generate a hybrid group of $G$ responses, partitioned into two subsets with equal size:
\vspace{-3mm}
\begin{itemize}[leftmargin=*]
\item \textbf{Exploitation Batch ($\mathcal{G}_{\text{std}}$):} Trajectories generated via standard sampling, which anchor the baseline to the policy's current bias.
\vspace{-3mm}
\item \textbf{Exploration Batch ($\mathcal{G}_{\text{CUTS}}$):} Trajectories generated via CUTS, which inject diversity by uncovering plausible but under-explored paths.\end{itemize}
\vspace{-3mm}

Advantages are computed over the combined group $\mathcal{G} = \mathcal{G}_{\text{std}} \cup \mathcal{G}_{\text{CUTS}}$. In saturated regimes where $\mathcal{G}_{\text{std}}$ collapses to uniform high rewards, $\mathcal{G}_{\text{CUTS}}$ introduces necessary outcome variability. This explicitly restores intra-group variance, generating informative relative signals for optimization.

\textbf{On-policy vs.\ Behavior Policy.}
Although Mixed-CUTS induces a mixed behavior policy $\mu$, we retain the standard clipped objective with $\pi_{\theta_{\text{old}}}$, and the resulting off-policy bias is tightly bounded by three compounding restrictions. \emph{First}, CUTS redistributes probability mass only within the Top-$K$ candidate set, so any token emitted by $\mu$ is already assigned non-trivial probability under the model's own distribution. \emph{Second}, the minimum-probability threshold $\delta$ prunes low-confidence tail tokens, keeping the proposal distribution inside a local trust region of semantically plausible continuations. \emph{Third}, the PPO clipping on the importance ratio $\pi_\theta/\pi_{\theta_{\text{old}}}$ further limits the per-step policy update, so even when a CUTS token has a low probability under $\pi_{\theta_{\text{old}}}$, the gradient contribution is clipped to a bounded range. Together these constraints ensure that Mixed-CUTS injects exploratory variance without severe divergence from the current policy, which is consistent with the empirically stable training curves observed across all of our runs.

\paragraph{Why Mixed-CUTS Restores the Advantage Signal.}
We now formalize why mixing an exploratory sub-group with the standard sub-group is guaranteed to keep the intra-group variance away from zero in saturated regimes. Consider a single prompt with a combined group $\mathcal{G}_{\text{mixed}}=\mathcal{G}_{\text{std}}\cup\mathcal{G}_{\text{CUTS}}$ of size $G$, split into two equal sub-groups of size $G/2$. Let $(\mu_{\text{std}},\sigma^2_{\text{std}})$ and $(\mu_{\text{CUTS}},\sigma^2_{\text{CUTS}})$ denote the sample mean and variance of the rewards within each sub-group. By the law of total variance applied to the combined group,
\begin{equation}
\label{eq:var_decomp}
\sigma^2_{\text{mixed}} = \tfrac{1}{2}(\sigma^2_{\text{std}}+\sigma^2_{\text{CUTS}}) + \tfrac{1}{4}(\mu_{\text{std}}-\mu_{\text{CUTS}})^{2}.
\end{equation}
The first ``within-group'' term captures the sampling noise inside each sub-group; the second ``between-group'' term is a non-negative penalty activated whenever the two sub-groups have different expected rewards. In a \emph{saturated} regime, standard GRPO corresponds to $\mathcal{G}=\mathcal{G}_{\text{std}}$ with $\sigma^2_{\text{std}}\to 0$, so the advantage in Eq.~\ref{eq:advantage} collapses. Because CUTS equalizes probabilities over the constrained Top-$K$ subset and therefore deviates from the greedy mode of $\pi_{\theta_{\text{old}}}$, it changes the expected per-prompt reward of the exploratory sub-group, i.e.\ $\mu_{\text{CUTS}}\neq\mu_{\text{std}}$: on ``too easy'' prompts ($\mu_{\text{std}}\to 1$) CUTS occasionally stumbles onto suboptimal branches, lowering $\mu_{\text{CUTS}}$; on ``too hard'' prompts ($\mu_{\text{std}}\to 0$) CUTS occasionally hits a correct alternative branch the greedy policy systematically misses, raising $\mu_{\text{CUTS}}$. Substituting either extreme into Eq.~\ref{eq:var_decomp} gives $\sigma^2_{\text{mixed}} \gtrsim \tfrac{1}{2}\sigma^2_{\text{CUTS}} + \tfrac{1}{4}(\mu_{\text{std}}-\mu_{\text{CUTS}})^{2} > 0$, so the intra-group variance is structurally prevented from collapsing as long as the exploratory sub-group behaves differently from the exploitative one. A complete case-by-case derivation of the two saturated extremes is provided in Appendix~\ref{app:variance_proof}.

\section{Experiments}

\begin{table*}[t]
\centering
\renewcommand{\arraystretch}{1.2}
\setlength{\tabcolsep}{4pt}
\caption{\textbf{Main results comparing \textsc{Mixed-CUTS} and GRPO on MATH.} We report Pass@1 and Pass@16 (\%) across five benchmarks, including "Thinking Mode" baselines for reference. $\Delta$ denotes the gain over GRPO.}
\vspace{-0.8em}
\resizebox{0.95\textwidth}{!}{
\begin{tabular}{ccccccccccc}
\toprule
\multirow{2}{*}{\textbf{Model}} & \multicolumn{2}{c}{\textbf{MATH}} & \multicolumn{2}{c}{\textbf{AIME24}} & \multicolumn{2}{c}{\textbf{AIME25}} & \multicolumn{2}{c}{\textbf{AMC}} & \multicolumn{2}{c}{\textbf{GPQA}} \\
\cmidrule(lr){2-3} \cmidrule(lr){4-5} \cmidrule(lr){6-7} \cmidrule(lr){8-9} \cmidrule(lr){10-11}
 & \textbf{Pass@1} & \textbf{Pass@16} & \textbf{Pass@1} & \textbf{Pass@16} & \textbf{Pass@1} & \textbf{Pass@16} & \textbf{Pass@1} & \textbf{Pass@16} & \textbf{Pass@1} & \textbf{Pass@16} \\
\midrule
\multicolumn{11}{c}{\textbf{Qwen3-1.7B(Non-thinking)}} \\
\midrule
Base Model  & 70.2 & 90.5 & 12.9 & 36.5 & 11.7 & 27.8 & 39.8 & 72.9 & 32.1 & 80.3 \\
GRPO & 83.6 & 93.9 & 29.5 & 60.7 & 22.8 & 44.5 & 59.8 & 82.9 & 34.2 & 80.1 \\
\sert{\textsc{Mixed-CUTS}} & \sert{85.1} & \sert{95.3} & \sert{32.3} & \sert{62.0} & \sert{28.1} & \sert{52.5} & \sert{62.7} & \sert{88.0} & \sert{36.0} & \sert{79.4} \\
$\Delta$ & \pos{1.5} & \pos{1.4} & \pos{2.8} & \pos{1.3} & \pos{5.3} & \pos{8.0} & \pos{2.9} & \pos{5.1} & \pos{1.8} & \negcolor{-0.7} \\
\midrule
Base Model (Thinking) & 82.6 & 93.2 & 28.9 & 62.8 & 24.9 & 44.5 & 57.5 & 82.0 & 34.9 & 75.3 \\
\midrule
\multicolumn{11}{c}{\textbf{Qwen3-4B (Non-thinking)}} \\
\midrule
Base Model & 82.5 & 94.4 & 24.2 & 54.0 & 21.5 & 48.0 & 61.3 & 82.9 & 45.3 & 83.9 \\
GRPO & 86.4 & 95.9 & 32.5 & 63.8 & 26.6 & 57.9 & 68.9 & 88.9 & 48.1 & 84.6 \\
\sert{\textsc{Mixed-CUTS}} & \sert{90.8} & \sert{96.6} & \sert{46.0} & \sert{73.5} & \sert{41.7} & \sert{71.9} & \sert{76.7} & \sert{91.9} & \sert{50.1} & \sert{84.5} \\
$\Delta$ & \pos{4.4} & \pos{0.7} & \pos{13.5} & \pos{9.7} & \pos{15.1} & \pos{14.0} & \pos{7.8} & \pos{3.0} & \pos{2.0} & \negcolor{-0.1} \\
\midrule
Base Model (Thinking) & 89.9 & 95.7 & 54.1 & 75.5 & 42.1 & 62.3 & 73.6 & 88.3 & 52.0 & 84.4 \\
\bottomrule
\end{tabular}
}
\vspace{-0.8em}
\label{tab:main}
\end{table*}

\begin{figure*}[t]
\centering
% TODO(camera-ready): regenerate Figure2.pdf to include a fourth subplot showing
% the AIME25 maj@16 training curves for standard GRPO vs. Mixed-CUTS.
% The caption below already references this fourth subplot.
\includegraphics[width=\textwidth]{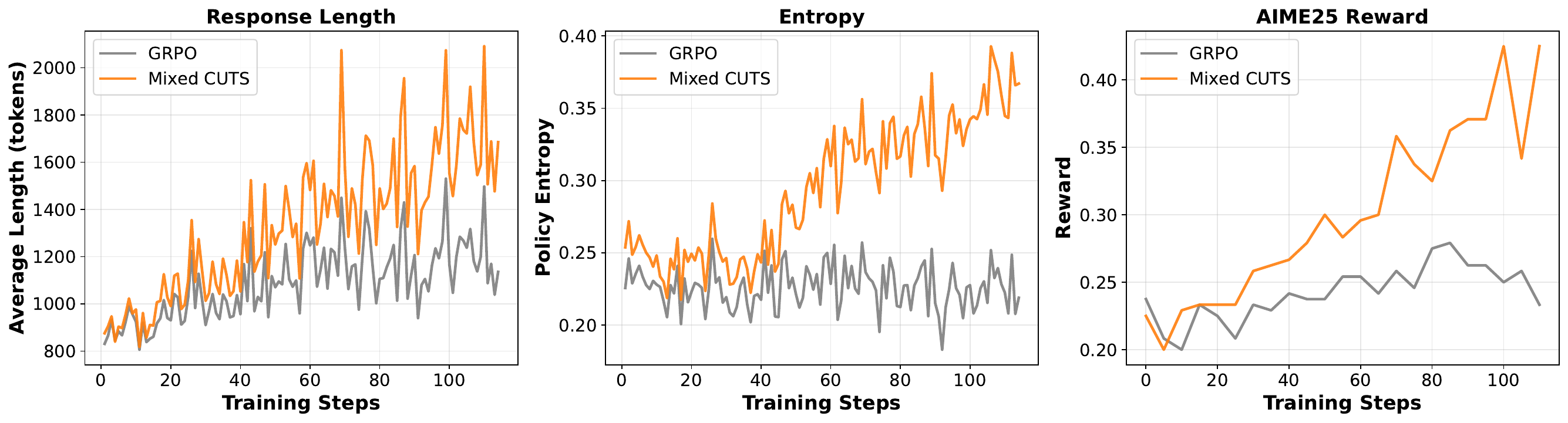}
\vspace{-8mm}
\caption{\textbf{Training Dynamics (Qwen3-4B).} Evolution of (Left) Response Length, (Middle-Left) Policy Entropy, (Middle-Right) AIME25 Reward, and (Right) AIME25 maj@16 consistency. Unlike GRPO (Grey), \textsc{Mixed-CUTS} (Orange) breaks the saturation trap by sustaining high entropy and inducing longer reasoning chains, driving both superior out-of-domain generalization and substantially stronger majority-vote consistency.}
\label{fig:dynamics}
\vspace{-4mm}
\end{figure*}

\subsection{Experimental Setup}
We train on the \textbf{MATH Training Set} \citep{hendrycks2021measuring} on Qwen3-1.7B and 4B (non-thinking mode). Details are in Appendix \ref{app:implementation_details}.

\subsection{Main Results}

Table \ref{tab:main} compares \textsc{Mixed-CUTS} with GRPO and "Thinking Mode" baselines. Results highlight four key insights:

\vspace{-2.5mm}\paragraph{Surpassing Intrinsic "Thinking" Capabilities.} \textsc{Mixed-CUTS} enables the 1.7B model to outperform its "Thinking Mode" (28.1\% vs 24.9\% on AIME25) without extended inference overhead. This suggests successful distillation of System-2-like reasoning into a standard, efficient policy.

\vspace{-2.5mm}\paragraph{Breaking the Saturation Trap.} \textsc{Mixed-CUTS} dominates out-of-domain. On Qwen3-4B, it beats GRPO by \textbf{+15.1\%} (AIME25) and \textbf{+13.5\%} (AIME24). This validates our saturation hypothesis: while GRPO collapses on easy data (MATH) due to vanishing gradients, our structured exploration sustains the advantage signal, driving robust generalization.

\vspace{-2.5mm}\paragraph{Robustness Over Randomness.} Pass@1 gains significantly outweigh Pass@16 gains (+4.4\% vs +0.7\% on 4B). This proves our method does not merely rely on random coverage but fundamentally shifts probability mass toward correct paths, improving policy reliability.

\vspace{-2.5mm}\paragraph{Scalability with Model Size.} Benefits amplify with scale (AIME25 gain: +5.3\% on 1.7B $\to$ +15.1\% on 4B). Larger models benefit more from CUTS's "width-first" exploration, which unlocks latent reasoning branches that standard greedy sampling prematurely prunes.

\subsection{Generalization beyond Mathematics}
\label{sec:cross_domain}
The gains of \textsc{Mixed-CUTS} are not limited to mathematical benchmarks. We evaluate our \emph{MATH-trained} Qwen3-4B checkpoints, without any further task-specific fine-tuning, on two comprehensive general-reasoning benchmarks outside the mathematical domain: \textbf{MMLU-Pro}, a multi-domain knowledge benchmark covering biology, law, literature, and more, and \textbf{SuperGPQA}, a graduate-level multi-discipline QA benchmark. As shown in Table~\ref{tab:cross_domain}, \textsc{Mixed-CUTS} consistently outperforms the standard GRPO baseline on both non-mathematical distributions (\textbf{+1.06\%} on MMLU-Pro and \textbf{+1.25\%} on SuperGPQA), despite the RL optimization being restricted to mathematical data. By preventing mode collapse on one axis (math), Mixed-CUTS enhances the model's broader structural exploration capabilities, transferring to diverse language tasks and confirming that our method improves \emph{fundamental} reasoning capabilities rather than overfitting to a single domain.

\begin{table}[h]
\small
\centering
\renewcommand{\arraystretch}{1.15}
\setlength{\tabcolsep}{6pt}
\caption{Zero-shot cross-domain accuracy of Qwen3-4B trained solely on MATH, evaluated on non-mathematical reasoning benchmarks.}
\label{tab:cross_domain}
\resizebox{\linewidth}{!}{
\begin{tabular}{lcc}
\toprule
\textbf{Training Method} & \textbf{MMLU-Pro} & \textbf{SuperGPQA} \\
\midrule
Base Model & 63.80\% & 33.05\% \\
Standard GRPO & 68.59\% & 40.03\% \\
\sert{\textsc{Mixed-CUTS} (Ours)} & \sert{\textbf{69.65\%}} & \sert{\textbf{41.28\%}} \\
\bottomrule
\end{tabular}
}
\end{table}

\subsection{Analysis of Training Dynamics}

Figure \ref{fig:dynamics} illustrates Qwen3-4B training dynamics, evidencing the efficacy of \textsc{Mixed-CUTS} against saturation.

\vspace{-2.5mm}\paragraph{Further Introducing Uncertainty.}The middle plot validates our mechanism. Standard GRPO (Grey) shows stagnant entropy ($\approx$0.20-0.25), confirming rapid convergence to "safe" patterns due to vanishing gradients. In contrast, \textsc{Mixed-CUTS} (Orange) sustains steady growth, acting as a variance-preservation mechanism that prevents premature convergence and keeps the optimization landscape active.\vspace{-2.5mm}\paragraph{Emergence of Deeper Reasoning.}Increased entropy signals deeper reasoning, not noise. While GRPO plateaus at $\approx$1200 tokens (Left), \textsc{Mixed-CUTS} drives trajectory lengths to peak over 1800. By forcing exploration of "second-best" tokens, CUTS unlocks latent "System 2" behaviors—such as self-correction—that are otherwise suppressed by greedy baselines.\vspace{-2.5mm}\paragraph{From Exploration to Generalization.}Extended exploration translates to robust generalization. On the harder AIME25 (Middle-Right), GRPO stagnates ($\approx$0.25), whereas \textsc{Mixed-CUTS} diverges sharply at step 30 to reach \textbf{0.40}. This correlation validates our premise: breaking the saturation trap on easy data can generalize LLMs to complex tasks.\vspace{-2.5mm}\paragraph{Consistency Gains Are a Stable Optimization Outcome.}The rightmost plot tracks the AIME25 maj@16 consistency metric over training. The $+23.2\%$ improvement reported in Table~\ref{tab:maj16_appendix} is not a late-stage artifact or random fluctuation: the \textsc{Mixed-CUTS} maj@16 curve pulls away from the GRPO curve early in training and the gap is robustly maintained throughout the later optimization phase, confirming that the consistency gain is a stable, continuously compounding outcome of the Mixed-CUTS objective rather than an artifact of the particular evaluation checkpoint.

\section{Conclusion}
We identified a critical bottleneck in LLM-RL: on saturated datasets, standard sampling induces vanishing gradients and mode collapse. To address this, we introduced \textbf{CUTS}, a lightweight operator enforcing local uniformity within the high-probability region of the model distribution. Integrated into the \textbf{Mixed-CUTS} framework, this approach explicitly restores informative advantage signals even in saturated regimes. Empirically, we confirm that such parameter-free diversification yields substantial gains in reasoning robustness and out-of-domain generalization. As models scale, strategies like CUTS that unlock latent capabilities beyond static data ceilings will be essential for continuous self-improvement. Future work will extend this uniform-prior exploration to code generation and agentic planning.

\subsection*{Limitation}
While Mixed-CUTS is motivated by the vanishing-advantage phenomenon in group-relative policy optimization, we do not yet provide a formal convergence or optimality analysis characterizing how decoding-time uniformization interacts with the GRPO objective. In particular, the mixed behavior policy induced by CUTS introduces a controlled deviation from strict on-policy sampling, and although this deviation is empirically stable under clipping, its long-horizon effect on policy improvement is not theoretically quantified. Moreover, our notion of diversity is operationalized through local distribution flattening within Top-$K$ candidates, which serves as a practical proxy rather than a formally grounded exploration criterion. Developing a principled theoretical framework that connects decoding-level interventions, advantage variance preservation, and convergence guarantees in saturated regimes remains an open direction for future research.

\bibliography{custom}

\newpage
\appendix

\section{Related Works} 
\textbf{RL for LLM Reasoning.} RL is the standard paradigm for enhancing LLM reasoning in objective domains like mathematics and coding \citep{liu2025deepseek,li2025self,wang2025causally,yu2025dapo,xiong2025rag,dai2025cde,liu2026save}. Recent advancements rely on GRPO and its variants \citep{guo2025deepseek, yang2025qwen3, yu2025dapo,wang2025beyond,wang2025adareasoner}, which efficiently estimate baselines via group averages without separate value networks. Despite this efficiency, optimization instability remains a challenge. Recent works observe that while RL elicits long chain-of-thought (CoT) reasoning, policies often rapidly converge to homogeneous generation patterns, stalling further improvement \citep{zhou2025evolving,liang2025can,yu2025restrain}.

\textbf{The Challenge of Saturation and Mode Collapse.} A critical bottleneck is the exploration-exploitation trade-off \citep{wang2025beyond,zhou2025evolving}. Traditionally, mode collapse was attributed to sparse rewards \citep{cui2025entropy,zhou2025dissecting}. However, we identify a distinct \textbf{saturation-induced collapse} in strong models on standard benchmarks. In these "easy-task" regimes, high baseline success rates cause intra-group reward variance to vanish \citep{yu2025dapo}. Lacking negative contrast, the relative advantage signal disappears, disincentivizing the exploration of superior strategies \citep{zhu2025surprising}. While entropy bonuses \citep{cui2025entropy} attempt to mitigate this, they often encourage nonsensical diversity: because entropy regularization indiscriminately penalizes confidence, it can disrupt coherent reasoning chains and lead to diversity that is incoherent rather than meaningful. In contrast, our decoding-time intervention (CUTS) performs \emph{structure-preserving} exploration: by restricting its uniform sampling strictly to the high-confidence Top-$K$ subset, CUTS maintains local semantic validity while effectively breaking the mode collapse, offering a more controlled and stable exploration mechanism than global entropy bonuses.

\section{Detailed Experimental Setup}
\label{app:implementation_details}

\subsection{Datasets} We conduct large-scale training using the canonical \textbf{MATH dataset} \citep{hendrycks2021measuring}. To rigorously evaluate our models, we employ a comprehensive suite of five benchmarks designed to measure both in-domain retention and out-of-domain generalization: \textbf{MATH-500}, \textbf{AIME24}, \textbf{AIME25}, \textbf{AMC} \citep{li2024numinamath}, and \textbf{GPQA-Diamond (GPQA)} \citep{rein2024gpqa}. Implementation details are provided in Appendix \ref{app:implementation_details}.

\subsection{Models and Configurations} We utilize the \textbf{Qwen3} series \citep{yang2025qwen3} as our backbone models, specifically the 1.7B and 4B parameter variants, with non-thinking mode. To fully accommodate the extensive reasoning chains required for complex mathematical problem solving, we scale the generation capacity according to the model size. We configure the maximum generation length to \textbf{5,000 tokens} for the 1.7B model and extend it to \textbf{12,000 tokens} for the 4B model. This extended context window is critical for avoiding truncation during the exploration of deep reasoning paths in the rollout phase.

% This section provides additional details on the implementation of our reward formulation and supporting mechanisms.

% \subsection{Training Configuration.} 

% We conduct our experiments on two recent open-source base models: \textbf{Qwen3-4B-Base} and \textbf{Qwen3-8B-Base}. Our training process is implemented using the GRPO algorithm. We adopt a setup similar to that of TTRL for generating training signals. For each problem instance, we first perform a rollout phase where the policy generates $64$ candidate responses. A majority label is then determined by performing a majority vote on the final answers extracted from these $64$ samples. Subsequently, a random subset of $32$ of these responses is used to form a batch for a single model update step. To ensure that the model has sufficient capacity for complex, multi-step reasoning, we set the maximum response length to 12,288 tokens during generation. To guide the model's reasoning process, we utilize the system prompt from SimpleRL-Zoo \citep{zeng2025simplerl}. Implementation details are discussed in Appendix \ref{app:implementation_details}.

\subsection{System Prompt}
For all experiments, we used the following system prompt to guide the model's generation format, ensuring that it produces a step-by-step reasoning process and a clearly marked final answer \citep{zeng2025simplerl}:
\begin{bluebox}[System Prompt]
\texttt{
Please reason step by step, and put your final answer within \textbackslash boxed\{\}.
}
\end{bluebox}
% \begin{verbatim}
% Please reason step by step, and put your final answer within \boxed{}.
% \end{verbatim}

\subsection{Answer and Reasoning Extraction}
To implement the scoring criteria described in the main text, we apply the following extraction procedure for each generated response $o_i$:
\begin{itemize}
    \item \textbf{Final Answer Extraction (for Validity):} We parse the response to find the content within the final occurrence of the \verb|\boxed{·}| command. A response is deemed "valid" only if this command is present and its content contains at least one numeric digit. This extracted numeric string is used for the majority vote.
\end{itemize}

\subsection{Hyperparameter Settings}
\label{app:hyperparameters}

This section summarizes the key hyperparameters used in \textsc{Mixed-CUTS} training and decoding. Unless otherwise specified, all parameters are fixed across experiments and model scales.

\paragraph{CUTS Decoding Hyperparameters.}
The CUTS operator is fully parameter-free with respect to model training and introduces only a small number of decoding-time hyperparameters. At each decoding step, we retrieve the Top-$K$ candidate tokens based on the model’s original distribution, with $K=5$ in all experiments. To ensure semantic plausibility, we apply a probability threshold $\delta=0.03$ to filter out low-confidence candidates. The remaining tokens are assigned a uniform probability, enforcing local width-first exploration. To preserve early reasoning stability, CUTS is disabled for the first $T_{\text{warmup}}=5$ tokens, during which standard decoding is applied.

\paragraph{Mixed-CUTS Sampling Strategy.}
Within the GRPO framework, we generate a group of $G=16$ trajectories per prompt. The group is evenly split into an exploitation batch ($G_{\text{std}}=8$), generated via standard sampling, and an exploration batch ($G_{\text{cuts}}=8$), generated using CUTS. Advantages are computed jointly over the combined group.

\paragraph{Training and Optimization Settings.}
We use a global training batch size of 128, with PPO mini-batches of size 32. KL regularization is enabled using a low-variance KL estimator with coefficient $1\times10^{-3}$. During rollout, we use temperature sampling with $T=1.0$ and disable nucleus truncation ($\text{top-}p=1.0$) to avoid confounding exploration effects. Validation rollouts follow standard decoding settings.

Table~\ref{tab:hyperparameters} provides a concise summary of the main hyperparameters.

\begin{table}[t]
\small
\centering
\renewcommand{\arraystretch}{1.1}
\setlength{\tabcolsep}{6pt}
\caption{Key hyperparameters used in \textsc{Mixed-CUTS}.}
\label{tab:hyperparameters}
\begin{tabular}{ll}
\toprule
\textbf{Category} & \textbf{Value} \\
\midrule
Top-$K$ ($K$) & 5 \\
Probability Threshold ($\delta$) & 0.03 \\
Warm-up Tokens ($T_{\text{warmup}}$) & 5 \\
Group Size ($G$) & 16 \\
Exploitation / Exploration Split & 8 / 8 \\
Training Batch Size & 128 \\
PPO Mini-batch Size & 32 \\
KL Loss Type & Low-variance KL \\
KL Coefficient & $1\times10^{-3}$ \\
Rollout Temperature & 1.0 \\
Rollout Top-$p$ & 1.0 \\
Validation Top-$p$ / Top-$k$ & 0.8 / 20 \\
Validation Temperature & 1.0 \\
\bottomrule
\end{tabular}
\end{table}

\subsection{Hyperparameter Sensitivity Analysis}
\label{app:hparam_sensitivity}

Beyond the default configuration in Table~\ref{tab:hyperparameters}, we independently probe the sensitivity of Mixed-CUTS to its two CUTS-specific decoding hyperparameters on AIME25, varying $\delta\in\{0.01,0.02,0.03,0.04,0.05\}$ at fixed $K=5$ and $K\in\{3,5,7,9\}$ at fixed $\delta=0.03$, for both Qwen3-1.7B and Qwen3-4B. Results are reported in Tables~\ref{tab:delta_sensitivity} and~\ref{tab:k_sensitivity}.

\begin{table}[h]
\small
\centering
\renewcommand{\arraystretch}{1.15}
\setlength{\tabcolsep}{4pt}
\caption{Impact of the minimum-probability threshold $\delta$ on AIME25 (fixed $K=5$). Pass@1 / Pass@16 (\%).}
\label{tab:delta_sensitivity}
\begin{tabular}{lcc}
\toprule
$\delta$ & \textbf{Qwen3-1.7B} & \textbf{Qwen3-4B} \\
\midrule
0.01 & 22.0 / 44.1 & 35.0 / 64.2 \\
0.02 & 26.5 / 50.2 & 40.0 / 69.5 \\
\sert{0.03 (Default)} & \sert{\textbf{28.1 / 52.5}} & \sert{\textbf{41.7 / 71.9}} \\
0.04 & 27.2 / 51.4 & 40.6 / 70.8 \\
0.05 & 26.0 / 49.8 & 39.8 / 70.1 \\
\bottomrule
\end{tabular}
\end{table}

\begin{table}[h]
\small
\centering
\renewcommand{\arraystretch}{1.15}
\setlength{\tabcolsep}{4pt}
\caption{Impact of the Top-$K$ candidate-set size on AIME25 (fixed $\delta=0.03$). Pass@1 / Pass@16 (\%).}
\label{tab:k_sensitivity}
\begin{tabular}{lcc}
\toprule
\textbf{Top-$K$} & \textbf{Qwen3-1.7B} & \textbf{Qwen3-4B} \\
\midrule
$K=3$ & 25.5 / 48.9 & 38.9 / 68.5 \\
\sert{$K=5$ (Default)} & \sert{\textbf{28.1 / 52.5}} & \sert{\textbf{41.7} / 71.9} \\
$K=7$ & 26.8 / 53.4 & 40.2 / \textbf{72.8} \\
$K=9$ & 21.5 / 48.2 & 35.5 / 68.2 \\
\bottomrule
\end{tabular}
\end{table}

\paragraph{Robustness across reasonable ranges.}
Performance remains highly stable across $\delta\in[0.02,0.05]$ and $K\in[3,7]$, and every configuration in these ranges consistently outperforms the standard GRPO baseline (26.6\% on Qwen3-4B AIME25), indicating that Mixed-CUTS does not require careful per-model tuning.

\paragraph{The necessity of the $\delta$ filter.}
Setting $\delta$ too low (e.g., $\delta=0.01$) causes a sharp drop (22.0\% on 1.7B, 35.0\% on 4B), because the weak filter allows low-quality tail tokens into the candidate set, and those tokens occasionally corrupt the reasoning chain. This empirically validates the design choice of filtering Top-$K$ by a minimum probability.

\paragraph{Exploration--noise tradeoff in $K$.}
$K=7$ achieves a slightly higher Pass@16 than the default $K=5$ on both model sizes, because a marginally wider search space uncovers more diverse correct paths when multiple samples are drawn. However, excessive $K$ ($K=9$) introduces noise, causing Pass@1 to drop much more severely (e.g., $-6.2\%$ on 4B) than Pass@16 ($-3.7\%$), matching the theoretical expectation that sampling noise heavily impacts single-shot accuracy but is partially mitigated by multi-sample evaluation.

\begin{table*}[t]
\centering
\renewcommand{\arraystretch}{1.2}
\setlength{\tabcolsep}{6pt}
\caption{Comparison of Majority Vote performance (maj@16) on the MATH dataset and out-of-domain benchmarks. Results represent the accuracy when selecting the most consistent answer from 16 sampled paths. $\Delta$ values indicate the improvement of Mixed-CUTS over the GRPO baseline.}
\vspace{-0.3em}
\resizebox{0.75\textwidth}{!}{
\begin{tabular}{lccccc}
\toprule
\textbf{Model} & \textbf{MATH} & \textbf{AIME24} & \textbf{AIME25} & \textbf{AMC} & \textbf{GPQA} \\
\midrule
\multicolumn{6}{c}{\textbf{Qwen3-1.7B}} \\
\midrule
Base Model (Non-thinking) & 77.6 & 21.1 & 16.2 & 50.9 & 34.3 \\
GRPO & 89.3 & 45.7 & 29.7 & 71.0 & 36.5 \\
\sert{\textsc{Mixed-CUTS}} & \sert{90.4} & \sert{49.2} & \sert{36.9} & \sert{74.3} & \sert{38.0} \\
$\Delta$ & \pos{1.1} & \pos{3.5} & \pos{7.2} & \pos{3.3} & \pos{1.5} \\
\midrule
Base Model (Thinking) & 88.9 & 45.4 & 31.3 & 68.6 & 38.7 \\
\midrule
\multicolumn{6}{c}{\textbf{Qwen3-4B}} \\
\midrule
Base Model (Non-thinking) & 88.3 & 33.1 & 24.8 & 70.5 & 48.1 \\
GRPO & 90.1 & 43.0 & 31.9 & 78.1 & 51.4 \\
\sert{\textsc{Mixed-CUTS}} & \sert{94.0} & \sert{54.9} & \sert{55.1} & \sert{83.0} & \sert{53.1} \\
$\Delta$ & \pos{3.9} & \pos{11.9} & \pos{23.2} & \pos{4.9} & \pos{1.7} \\
\midrule
Base Model (Thinking) & 92.4 & 67.5 & 54.0 & 81.2 & 54.8 \\
\bottomrule
\end{tabular}
}
\label{tab:maj16_appendix}
\end{table*}

\section{Additional Experiments}

\subsection{Robustness Analysis: Majority Vote Performance}

In addition to Pass@1 and Pass@16, we evaluate the models using Majority Vote (maj@16), a metric that reflects the model's internal consistency and confidence. Unlike Pass@N, which measures the existence of a correct solution in the sample space, maj@16 measures whether the correct solution dominates the probability distribution. The results are detailed in Table \ref{tab:maj16_appendix}.

\paragraph{Significant Gains in Solution Consistency.} \textsc{Mixed-CUTS} demonstrates remarkable improvements in consistency compared to the GRPO baseline. On the Qwen3-4B model, we observe a massive +23.2\% improvement on AIME25 (maj@16 increases from 31.9\% to 55.1\%). This indicates that our method does not simply "stumble upon" the correct answer through random exploration; rather, it fundamentally reshapes the policy to assign high probability mass to correct reasoning paths. Standard GRPO, by contrast, often struggles to achieve consensus on hard tasks due to optimization instability, leading to lower majority vote scores despite decent Pass@N performance.

\paragraph{Beating "Thinking Mode" Consistency.} It is particularly noteworthy that \textsc{Mixed-CUTS} (operating in standard mode) achieves higher consistency than the base model's native "Thinking Mode" on several key benchmarks. For instance, on the 4B scale, \textsc{Mixed-CUTS} achieves a maj@16 of 55.1\% on AIME25, surpassing the Thinking Mode's 54.0\%. Similarly, on the 1.7B scale, our method outperforms Thinking Mode on AIME25 (36.9\% vs 31.3\%) and AMC (74.3\% vs 68.6\%). This result reinforces our claim that \textsc{Mixed-CUTS} effectively distills the benefits of extensive search into a robust, low-latency policy that yields reliable, consistent solutions without the computational overhead of recursive thinking.

\subsection{Performance with Abundant Hard Data}
\label{app:dapo}

We further verify that Mixed-CUTS remains effective when trained directly on high-quality hard data, showing that its benefits are orthogonal to data difficulty rather than a substitute for hard data. We train Qwen3-4B directly on the \textbf{DAPO-17K} dataset, a large-scale collection of high-quality reasoning prompts analogous in difficulty to AIME-level training distributions, and evaluate on the same five benchmarks used in the main text (Table~\ref{tab:dapo}).

\begin{table*}[h]
\small
\centering
\renewcommand{\arraystretch}{1.15}
\setlength{\tabcolsep}{5pt}
\caption{Performance of Qwen3-4B trained directly on the hard \textbf{DAPO-17K} dataset. Pass@1 / Pass@16 (\%) across five reasoning benchmarks.}
\label{tab:dapo}
\resizebox{\textwidth}{!}{
\begin{tabular}{lccccc}
\toprule
\textbf{Model (Qwen3-4B, trained on DAPO-17K)} & \textbf{AIME24} & \textbf{MATH-500} & \textbf{AIME25} & \textbf{AMC} & \textbf{GPQA} \\
\midrule
Standard GRPO & 65.6 / 82.6 & 92.6 / 96.3 & 54.1 / 68.7 & 84.6 / 95.6 & 54.9 / 81.9 \\
\sert{\textsc{Mixed-CUTS} (Ours)} & \sert{\textbf{67.5 / 84.1}} & \sert{\textbf{93.8 / 97.8}} & \sert{\textbf{56.0 / 73.6}} & \sert{\textbf{87.7 / 97.7}} & \sert{\textbf{56.6 / 88.9}} \\
\bottomrule
\end{tabular}
}
\end{table*}

\paragraph{Consistent gains on hard data.}
Training on harder data raises the performance floor: standard GRPO on DAPO reaches 54.1\% on AIME25 Pass@1, compared to 26.6\% on MATH. On top of this stronger baseline, \textsc{Mixed-CUTS} still yields consistent absolute gains across all five benchmarks (e.g., $+1.9\%$ on AIME25, $+3.1\%$ on AMC, $+1.7\%$ on GPQA for Pass@1), with even larger margins on Pass@16 (e.g., $+4.9\%$ on AIME25, $+7.0\%$ on GPQA). The ``vanishing advantage'' phenomenon re-emerges once the model begins to saturate even on harder data, and Mixed-CUTS continues to break this new saturation bottleneck by systematically exploring valid alternative semantic branches.

\paragraph{Beyond the ``data wall''.}
The absolute algorithmic gain is larger on MATH than on DAPO ($+15.1\%$ vs.\ $+1.9\%$ on AIME25 Pass@1), and this pattern highlights where Mixed-CUTS contributes most. When abundant, ultra-hard labeled data is available, the inherent difficulty of the prompts already forces the model into high-variance exploration, so the vanishing-advantage collapse is less severe. The more interesting regime is the opposite one: curating increasingly difficult high-quality reasoning datasets becomes unsustainably expensive as model capabilities scale. Mixed-CUTS shows that \emph{easy}, easily-saturated data still carries exploitable learning signal---by structurally enforcing exploration on simple datasets like MATH, the model acquires generalized reasoning skills (the $+15.1\%$ gain on AIME25) without needing an endless supply of DAPO-level prompts. Even when hard data \emph{is} abundant, Mixed-CUTS still provides orthogonal gains on top of it.

\section{Variance Preservation: Full Derivation}
\label{app:variance_proof}

This appendix provides the complete case-by-case derivation of the variance-preservation argument sketched in Section~\ref{sec:method} (the discussion following Eq.~\ref{eq:var_decomp}). The goal is to show that, in the two saturated extremes that kill the GRPO advantage signal, the intra-group variance $\sigma^2_{\text{mixed}}$ of a Mixed-CUTS group is strictly bounded away from zero.

\paragraph{Setup.}
For a single prompt $\mathbf{q}$, let $\mathcal{G}_{\text{std}}$ and $\mathcal{G}_{\text{CUTS}}$ be the standard and CUTS sub-groups of size $G/2$, with binary rewards $r_i\in\{0,1\}$ (correct / incorrect). Let $(\mu_{\text{std}},\sigma^2_{\text{std}})$ and $(\mu_{\text{CUTS}},\sigma^2_{\text{CUTS}})$ be their sample means and variances. The combined group has size $G$, mean $\mu_{\text{mixed}}=\tfrac{1}{2}(\mu_{\text{std}}+\mu_{\text{CUTS}})$, and variance
\[
\sigma^2_{\text{mixed}} = \tfrac{1}{2}(\sigma^2_{\text{std}}+\sigma^2_{\text{CUTS}}) + \tfrac{1}{4}(\mu_{\text{std}}-\mu_{\text{CUTS}})^{2},
\]
by the law of total variance for two equal sub-groups.

\paragraph{Key behavioral assumption.}
By construction, CUTS does not follow the model's greedy mode: within the Top-$K$ subset filtered by the probability threshold $\delta$, it replaces the model's skewed distribution with a uniform one. Whenever $|\mathcal{S}_t|\ge 2$ and the Top-$K$ mass is non-trivially concentrated on the greedy token, the exploratory sub-group therefore produces trajectories that differ semantically from the greedy trajectories generated by $\pi_{\theta_{\text{old}}}$. In expectation over prompts, this behavioral gap implies $\mu_{\text{CUTS}}\neq\mu_{\text{std}}$ on any prompt where the model is not already deterministic.

\paragraph{Case A: ``Too easy'' saturated prompt ($\mu_{\text{std}}\to 1$, $\sigma^2_{\text{std}}\to 0$).}
All standard samples succeed, so the greedy policy is essentially deterministic on this prompt and the within-group variance of $\mathcal{G}_{\text{std}}$ vanishes. CUTS, by decoupling sampling from the peaked distribution, occasionally selects a semantically valid but sub-optimal branch that does not lead to the canonical solution; some of these branches fail, pushing $\mu_{\text{CUTS}}$ below $1$. Substituting $\mu_{\text{std}}\to 1$, $\sigma^2_{\text{std}}\to 0$, $\mu_{\text{CUTS}}<1$ into the decomposition yields
\[
\sigma^2_{\text{mixed}} \approx \tfrac{1}{2}\sigma^2_{\text{CUTS}} + \tfrac{1}{4}(1-\mu_{\text{CUTS}})^{2} > 0.
\]
A non-zero advantage signal is therefore preserved for exactly the ``all-correct'' prompts where standard GRPO breaks down.

\paragraph{Case B: ``Too hard'' saturated prompt ($\mu_{\text{std}}\to 0$, $\sigma^2_{\text{std}}\to 0$).}
All standard samples fail because the greedy policy repeatedly commits to the same incorrect reasoning path. CUTS forces the model to uniformly consider its Top-$K$ alternatives, giving it a non-trivial probability of stumbling onto a correct intermediate step that the greedy mode systematically misses; some of these exploratory trajectories succeed, pushing $\mu_{\text{CUTS}}$ above $0$. Substituting $\mu_{\text{std}}\to 0$, $\sigma^2_{\text{std}}\to 0$, $\mu_{\text{CUTS}}>0$ into the decomposition yields
\[
\sigma^2_{\text{mixed}} \approx \tfrac{1}{2}\sigma^2_{\text{CUTS}} + \tfrac{1}{4}\mu_{\text{CUTS}}^{2} > 0.
\]
Even on prompts where standard GRPO sees only failures, Mixed-CUTS recovers a positive variance and, importantly, a positive advantage $\hat{A}_i$ for the rare CUTS trajectories that happen to succeed---exactly the learning signal required to escape the ``too-hard'' failure mode.

\paragraph{Conclusion.}
In both saturated extremes, $\sigma^2_{\text{mixed}}$ is strictly lower-bounded by a non-zero quantity driven by the between-group difference $(\mu_{\text{std}}-\mu_{\text{CUTS}})^{2}$. The standardized advantage in Eq.~\ref{eq:advantage} is therefore kept away from the degenerate $0/\epsilon$ regime, and the policy gradient remains informative. This is the formal statement of the ``structural variance preservation'' claim made in the main text: Mixed-CUTS does not rely on noise to restore contrast---it relies on the structural behavioral gap between greedy and Top-$K$-uniform decoding, and this gap is precisely the second term of Eq.~\ref{eq:var_decomp}.

\section{AI Writing Assistance Declaration}
We utilized generative AI models solely to improve the readability and clarity of the manuscript. The scope of assistance was limited to grammatical correction and stylistic polishing of the content originally written by the authors.

\end{document}